\documentclass[preprint,12pt]{elsarticle}
\usepackage{amssymb}
\usepackage{amsmath}
\usepackage{multirow}
\usepackage{subcaption}
\usepackage{adjustbox}

\makeatletter
\def\ps@pprintTitle{%
  \let\@oddhead\@empty
  \let\@evenhead\@empty
  \def\@oddfoot{\centerline{\thepage}}%
  \let\@evenfoot\@oddfoot}
\makeatother

\begin{document}

\begin{frontmatter}

\title{AI-Enhanced Hyperspectral Imaging for Intraoperative Detection of Low-Grade Glioma: A Clinical Study}

\author[label1]{Siri Willems}
\author[label1,label2]{Ishika Jain}
\author[label1,label6]{Jens De Winne}
\author[label1]{Siri Luthman}
\author[label1]{Hamed Zivari Adab}
\author[label1]{Roeland Vandebriel}
\author[label4]{Anna Alperovich}
\author[label5]{Xiaohan Zhang}
\author[label1]{Ben Stoffelen}
\author[label1]{Tom De Schepper}
\author[label3]{Lien Solie}
\author[label3]{Steven De Vleeschouwer}

\affiliation[label1]{organization={imec},
            addressline={Kapeldreef 75}, 
            city={Leuven},
            postcode={3001}, 
            country={Belgium}}

\affiliation[label2]{organization={Department of Computer Science UAntwerp-imec},
            addressline={Sint-Pietersvliet 7}, 
            city={Antwerp},
            postcode={2000},
            country={Belgium}}

\affiliation[label6]{organization={Department of Telecommunications and Information Processing, IPI-imec},
            addressline={Sint-Pietersnieuwestraat 41}, 
            city={Ghent},
            postcode={9000}, 
            country={Belgium}}
            
\affiliation[label3]{organization={Research Group Experimental Neurosurgery and Neuroanatomy, Dept Neurosciences, KU Leuven, Leuven Brain Institute and Dept. Neurosurgery, UZ Leuven},
            addressline={Herestraat 49}, 
            city={Leuven},
            postcode={3000},
            country={Belgium}}

\affiliation[label4]{organization={Carl Zeiss AG, Corporate Research \& Technology},
            addressline={Carl-Zeiss-Str. 22}, 
            city={Oberkochen},
            postcode={73447},
            country={Germany}}

\affiliation[label5]{organization={Carl Zeiss Meditec AG},
            addressline={Rudolf-Eber-Str. 11}, 
            city={Oberkochen},
            postcode={73447},
            country={Germany}}

\begin{abstract}
Current image-guided neuronavigation techniques, such as intraoperative MRI and ultrasound, are limited by high costs, limited sensitivity and operational complexity. There is a clear need for more efficient systems to enhance intra-operative tumor detection and improve patient outcomes for intrinsic brain tumor resection. Whereas previous intraoperative brain HSI studies have predominantly investigated high-grade gliomas, glioblastomas, or mixed-grade cohorts, this clinical study focuses specifically on low-grade gliomas (LGGs). LGGs strongly resemble normal brain under white light conditions on the operative microscope. This study characterizes spectral differences between healthy and tumor tissue and provides a paired comparison of HSI- and RGB-based classification at matched tissue locations and under identical patient-level data splits. Tumor regions showed consistently lower reflectance, which may be associated with differences in hemoglobin concentration and oxygenation. A range of classification models was evaluated. The 3D CNN, which integrates spectral and spatial features, achieved the highest median ROC-AUC of 74.5\%, compared with 67.9\% for the spectral-only XGBoost model and 64.0\% for the RGB-based model. These results support further investigation of HSI for low-grade glioma classification and quantify its performance relative to conventional RGB imaging in this cohort.
\end{abstract}

\end{frontmatter}

\section{Introduction}
Brain and central nervous system (CNS) cancers are among the most fatal cancers worldwide, and their incidence is rising globally \cite{noauthor_global_nodate}. These include primary brain tumors, i.e. an irregular growth of malignant cells starting from intrinsic brain cells, and secondary tumors that metastasize to the brain from elsewhere in the body. Gliomas are a major group of primary brain tumors whose diagnosis and grading under the 2021 WHO classification depend on histological and molecular features and the presence or absence of an IDH mutation \cite{louis_2021_2021}. Different groups are low-grade gliomas (LGG, grade 1 and 2), high-grade gliomas (HGG, grade 3) and glioblastomas (GBM, grade 4). Surgical resection is the first step of the treatment, but achieving an optimal balance between complete tumor removal and preservation of healthy functional tissue remains challenging. Maximizing the extent of resection can improve patient survival, but resection near eloquent tissue may impair neurological function and postoperative quality of life \cite{sanai_glioma_2008, stummer_counterbalancing_2011}. Distinguishing low-grade gliomas (LGGs) from healthy tissue intra-operatively is particularly difficult because LGGs can closely resemble healthy tissue, especially at the tumor periphery \cite{leon_hyperspectral_2023,jakola_surgical_2017}.\par 

Neurosurgery relies heavily on image-guided neuronavigation, but current methods have several limitations. Stereotactic neuronavigation uses preoperative magnetic resonance imaging (MRI) and a predefined surgical plan, and its accuracy decreases as brain shift occurs after craniotomy. Intraoperative MRI can visualize current surgical anatomy but is expensive, logistically complex, and time-consuming. Intraoperative ultrasound (US) is faster and less expensive but can be difficult to interpret and is affected by artifacts from bone and blood. Fluorescence imaging requires administration of a fluorophore and is primarily effective for high-grade gliomas, because most LGGs do not fluoresce. More efficient intraoperative guidance systems could therefore complement current techniques \cite{leon_hyperspectral_2023,gerard_brain_2017,lakomkin_fluorescence-guided_2018,sastry_applications_2017, verburg_state---art_2021,zhang_role_2023, barberio_intraoperative_2021, leroy_high-field_2019,bo_intraoperative_2020, stummer_fluorescence-guided_2006,almekkawi_use_2020}.\par

Hyperspectral imaging (HSI) combines conventional imaging with spectroscopy. Unlike RGB cameras, which capture three broad color bands, HSI systems acquire many distinct, contiguous bands across an instrument's spectral range, from ultraviolet to short-wave infrared depending on the device \cite{karim2023hyperspectral, noauthor_ieee_nodate}. The resulting hypercube contains spatial and spectral information that enables detailed tissue analysis \cite{jacques_optical_2013, halicek_-vivo_2019}. Medical interest in HSI has increased because it can enhance fluorescent biomarkers \cite{black_towards_2024} and provide rich spectral information without exogenous labels. HSI may support applications in gastroenterology, neuro-oncology, histopathology, and other medical fields \cite{lu_medical_2014, fei_hyperspectral_2020, clancy_surgical_2020, shapey_intraoperative_2019,karim2023hyperspectral}.\par

Due to the limited human interpretability of hyperspectral data, interest in AI-based analysis of hyperspectral images has increased. The intraoperative neuro-oncology studies reviewed here predominantly address glioblastoma, high-grade glioma, or cohorts containing multiple tumor grades, often through pixel-wise classification of tumor and healthy brain tissue. Martin-Perez et al. compared snapshot and linescan HSI cameras using a pixel-wise random forest classifier \cite{martin-perez_spectral_2024}. Fabelo et al. proposed a hybrid framework combining a U-Net for parenchymal masking, a 2D CNN for spatial analysis and vessel identification, and a 1D CNN for tissue characterization \cite{fabelo_deep_nodate}. Urbanos et al. evaluated a 3D CNN that jointly captures spectral and spatial features for pixel-wise glioblastoma classification and compared it with support-vector-machine and random-forest classifiers \cite{urbanos_supervised_2021}. Puustinen et al. developed ML and DL methods for high-grade glioma detection \cite{puustinen_hyperspectral_2023}. Manni et al. combined PCA-based spectral dimensionality reduction, 3D convolutional layers for spatio-spectral learning, and 2D convolutional layers for spatial learning \cite{manni_hyperspectral_2020}. In contrast to this emphasis on high-grade, glioblastoma, or mixed-grade populations, the present study evaluates a clinical cohort focused on low-grade glioma.\par

In this paper, we present the results of a clinical study in which 18 patients were enrolled to investigate the differentiation of low-grade glioma from healthy tissue using HSI. Of these patients, 17 were retained for model development after one patient was excluded because fluorescence imaging was used. Relative to the intraoperative brain HSI studies reviewed above, which predominantly investigated high-grade glioma, glioblastoma, or mixed-grade cohorts, the distinguishing focus of this study is a low-grade glioma cohort, which is the most challenging glioma entity in terms of intraoperative tumor detection. Preliminary results from five patients were published in a conference proceeding \cite{giannantonio_intra-operative_2023}. The study has two main aims:

\begin{itemize}
\item First, we compare this low-grade glioma dataset with the publicly available HELiCoiD dataset, which contains multiple tumor grades. We assess inter-patient and intra-patient spectral variability for different tissue types. HELiCoiD has previously been used to benchmark spectral imaging datasets \cite{martin2025unifying}. This comparison contributes to the standardized assessment of spectral imaging for surgical guidance, particularly for low-grade glioma.

\item Second, we examine the difference between HSI and conventional RGB imaging in greater depth through a paired model comparison. Multiple machine-learning and deep-learning methods are trained on the internal HSI data and compared with a model trained on RGB images acquired from the same tissue regions with a standard surgical microscope. All models use the same patient-level splits and evaluation procedure, allowing the contribution of HSI information to be assessed under matched conditions.
\end{itemize}

\section{Data}
Two datasets are used; their camera and acquisition parameters are summarized in Table \ref{tab:acquisition_params}. The first is a public dataset acquired during the European HypErspectraL Imaging Cancer Detection (HELiCoiD) project \cite{leon_hyperspectral_2023, fabelo_-vivo_2019}, and the second is the internal InVivo-LGG dataset acquired during our clinical study \cite{vandebriel_integrating_2023}.\par

\begin{table}[h]
\centering
\caption{Overview of dataset characteristics for both internal and public datasets used in the study.}\label{tab:acquisition_params}
\begin{tabular}{|p{4cm}| p{4.5cm}| p{4.5cm}|}
\hline
\textbf{Specifications}&	\textbf{InVivo-LGG}  &	\textbf{HELiCoiD} \cite{leon_hyperspectral_2023}\\
\hline
\hline
Tumor types &	LGG	& Mix of brain tumors\\
\hline
Number of patients	&18&	34\\
\hline
Number of hypercubes&	75&	61\\
\hline
Maximum spatial resolution in pixels&	$1600\times1600$	&$741\times1004$\\
\hline
Spectral range&	470--787 nm	&400--1000 nm\\
\hline
Spectral resolution&	10--15 nm	&2--3 nm\\
\hline
Number of bands	&104	&826\\
\hline
HSI camera&	imec Snapscan VNIR&	Hyperspec VNIR A-Series (Headwall Photonics Inc.)\\
\hline
Light source&	Xenon	&Quartz tungsten halogen\\
\hline
White reference&	Zenith Polymer (95\% reflectance)	& White tile (99\% reflectance)\\
\hline
Acquisition time	&120 s&	60 s\\
\hline
\end{tabular}

\end{table}

\subsection{InVivo-LGG dataset (internal, UZ Leuven)}
\subsubsection{Hardware setup and data acquisition}
\begin{figure*}[h]
    \centering
    \includegraphics[width=1\textwidth]{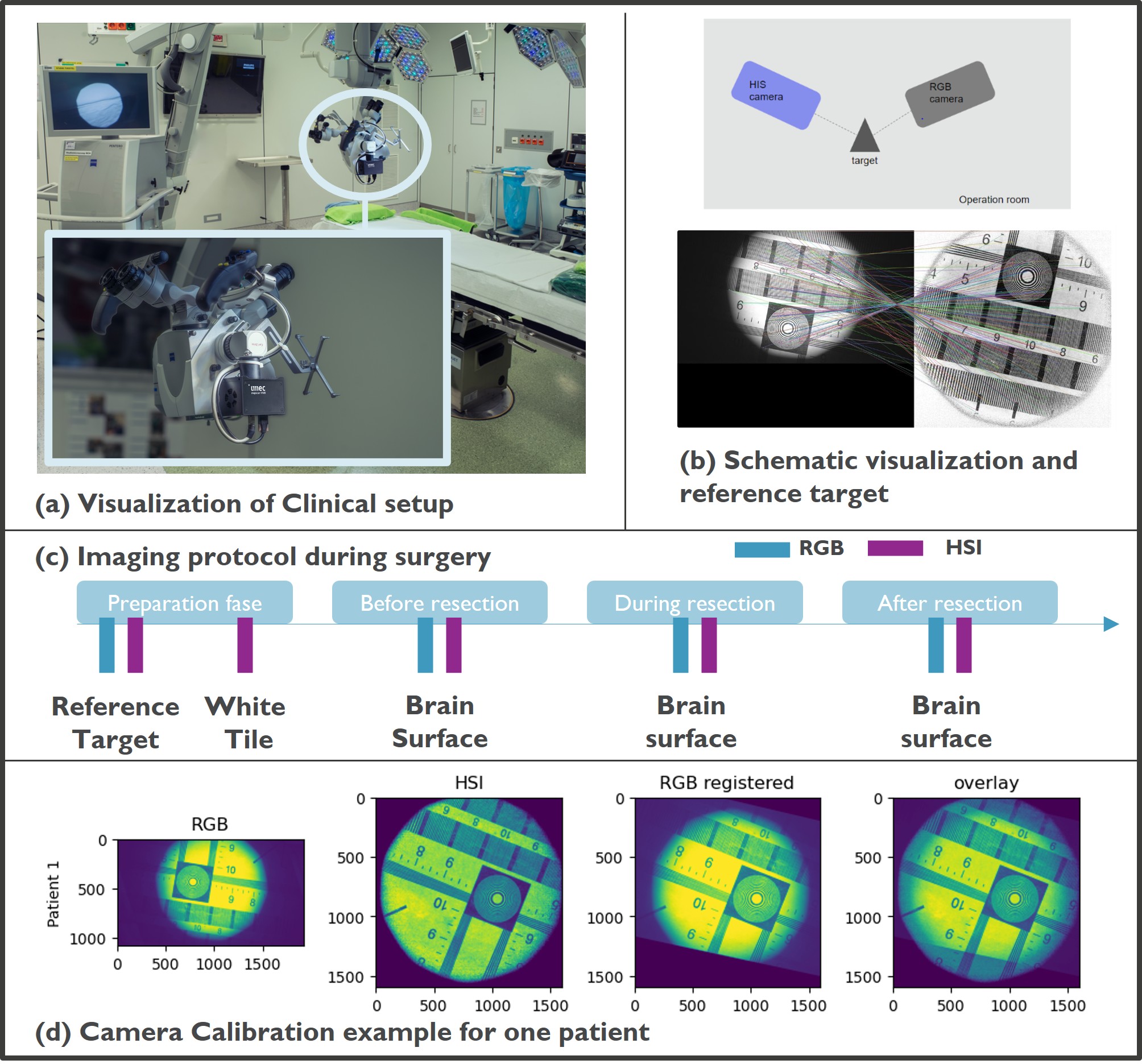}
    \caption{(a) Clinical setup. (b) Reference target used to align the RGB and HSI cameras. (c) Image-acquisition protocol. (d) RGB--HSI calibration results.}
    \label{fig:clinicalsetup}
\end{figure*}

The setup was previously described by Vandebriel et al. \cite{vandebriel_integrating_2023} and is summarized here. An intraoperative imaging system was installed at UZ Leuven by attaching an imec Snapscan VNIR hyperspectral camera (imec, Leuven, Belgium) to the lateral co-observer of a ZEISS OPMI Pentero 900 neurosurgical microscope using a standard camera-port adapter (ZEISS camera adapter, $f=105$ mm). The microscope's optics and illumination channel were used for HSI acquisition. The microscope was operated at a focal distance of 250 mm and a zoom factor of 2.5 to match a typical standard-of-care field of view (Figure \ref{fig:clinicalsetup}).\par
During each surgery, a predefined protocol was followed (Figure \ref{fig:clinicalsetup}c). First, a spatial reference target printed on A4 paper was imaged with both the microscope's integrated RGB camera and the attached HSI camera for subsequent spatial calibration. Second, a white-reference image was acquired by scanning a Zenith Polymer target (95\% reflectance; SphereOptics, Germany) with the HSI camera before surgery. Images of the brain surface were acquired before, during, and after resection. To limit temporal changes, paired images were acquired sequentially: an RGB image was followed immediately by an HSI scan of the same region of interest \cite{verburg_state---art_2021, zhang_applications_2020}.\par

A clinical study (ID: NCT04859725 [33]) was conducted in which 18 patients were enrolled after providing written informed consent. The imaging protocol resulted in 75 intraoperative brain images. The head surgeon (S. DV), who performed all surgeries, annotated the RGB images directly after surgery using anatomical references from neuronavigational MRI, intraoperative US, and biopsies obtained at selected locations. The following classes were annotated: 1: "Healthy", 2: "Foreign object", 3: "Blood", 4: "Coagulation", 5: "HGG", 6: "LGG", 7: "Histology HGG", and 8: "Histology LGG". Unlabeled pixels were considered background. Patient 10 received fluorescence imaging and was excluded from model development and evaluation, leaving 17 patients in the modeling cohort. Patient 3 contained only healthy-tissue annotations and was retained for model training but excluded from patient-level binary classification evaluation, which requires both classes. Consequently, 16 patients contributed to the patient-level results.\par

\subsubsection{Data preprocessing}
As described by Vandebriel et al. \cite{vandebriel_integrating_2023}, the spectral images were dark-corrected, white-balanced, and spectrally corrected. Dark correction and white balancing reduce sensor- and system-level noise. Spectral correction suppresses variation in the manufactured spectral bands by constructing idealized responses from linear combinations of the raw band responses. Bands with peak responses outside the microscope light source's spectral range were discarded. The resulting data contain 104 corrected contiguous bands spanning 470--787 nm.\par 

Annotations were created in the RGB frame of reference. To preserve the HSI data, the RGB images and corresponding delineations were registered to the HSI frame using a two-step procedure. Reference-target images were used for initial camera calibration with the scale-invariant feature transform in OpenCV. The resulting transformation was applied to the brain-surface RGB images and delineations. To compensate for possible camera movement during surgery, a second rigid registration was performed with SimpleElastix, allowing only small rotations and translations. All registrations were visually verified by overlaying the registered RGB images and delineations on the HSI images; examples are provided in Figure \ref{fig:registrationexamples}. Misaligned cases were manually registered with the open-source ITK-SNAP software. Because illumination falloff and optical aberrations degrade image quality near the circular field-of-view boundary, all 104-band hypercubes were cropped to $800\times800$ pixels around the center. Figure \ref{fig:visualization}(a) shows an RGB image, the mean HSI image across spectral bands, and the corresponding ground-truth labels.\par

\subsubsection{Ground truth processing}
The manual ground-truth annotations registered to the HSI images include small vascular structures, specular reflections, and edge noise. Additional processing was therefore applied to remove outliers. First, each HSI image was divided into spatially contiguous superpixels with similar spectral features. A circular mask with a diameter of 800 pixels was created, and simple linear iterative clustering (SLIC) was used to generate superpixels with a target area of 200 pixels \cite{achanta_slic_2012}.\par

\begin{figure}[h]
    \centering
    \includegraphics[width=1\textwidth]{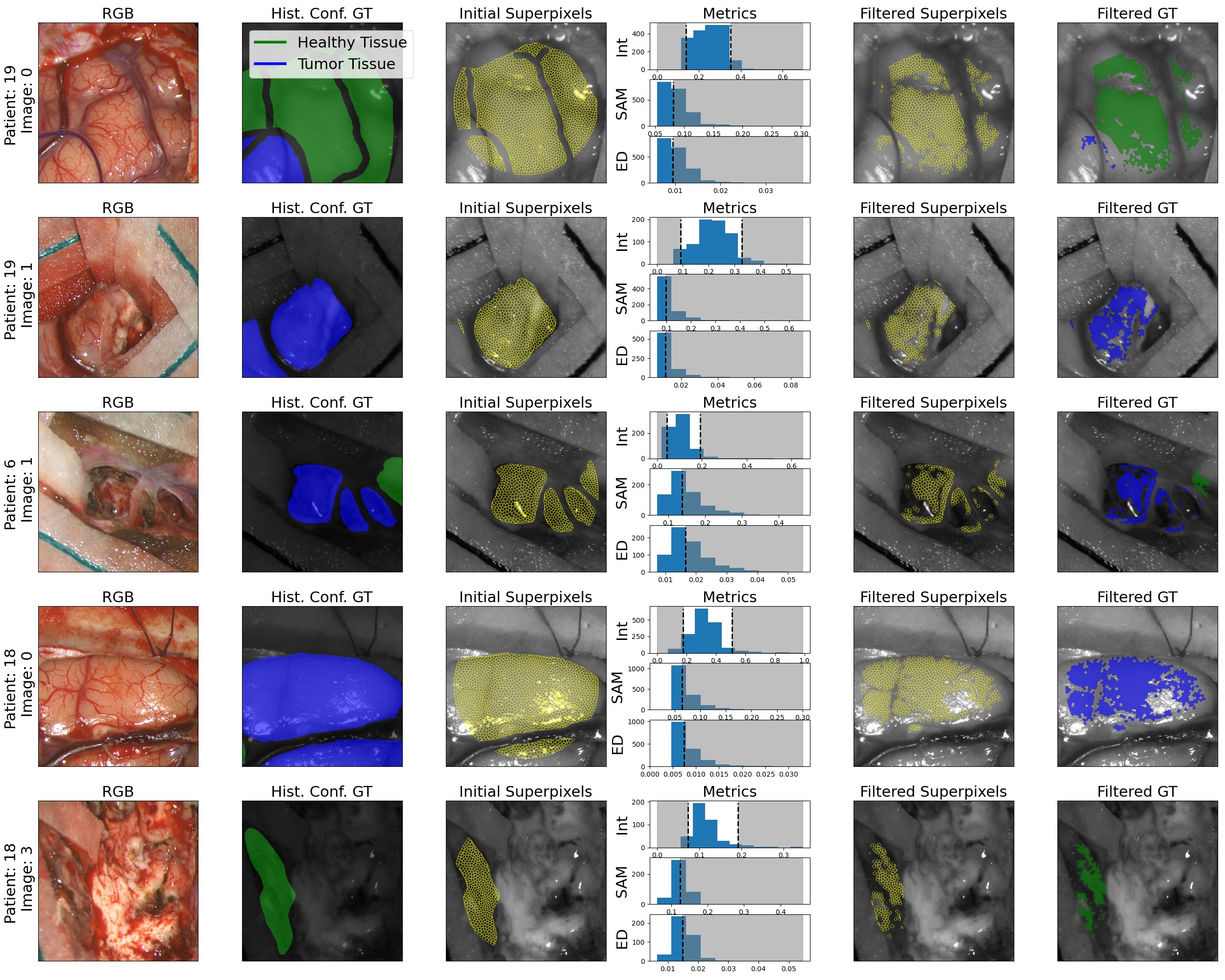}
    \caption{RGB image, histology-confirmed ground truth, initial superpixels, metric distributions, filtered superpixels, and final filtered ground truth.}
    \label{fig:visualization}
\end{figure} 

To improve delineation quality and remove outliers, including specular reflections, vessels annotated as tissue, and noisy superpixels, variability within and between superpixels was evaluated using the following metrics:\par

First, only superpixels that contain histology confirmed ground truths for healthy and LGG tumor tissue are kept. If a superpixel contains a mix of different labels, the superpixel is removed.\\
Second, the average intensity I$_S$ within the superpixel is calculated as:\\
\begin{equation}
I_s = \frac{1}{n}\sum{I_p}
\end{equation}
with I$_p$ being the intensity of each individual pixel, belonging to the superpixel.
Superpixels with intensities outside the image-specific 5th--95th percentile range, which primarily represent dark vessels or bright specular reflections, were excluded (Figure \ref{fig:visualization}b).\\

Third, Euclidean distance (ED) was used to quantify spectral-shape variability within each superpixel. Each pixel spectrum $\mathbf{f}_p$ was first L2-normalized to remove multiplicative magnitude differences:
\begin{equation}
\tilde{\mathbf{f}}_p =
\frac{\mathbf{f}_p}{\left\|\mathbf{f}_p\right\|_2}.
\end{equation}
The mean normalized spectrum of superpixel $s$ was then calculated as
\begin{equation}
\bar{\mathbf{f}}_s = \frac{1}{n_s}
\sum_{p=1}^{n_s}\tilde{\mathbf{f}}_p,
\end{equation}
where $n_s$ is the number of pixels in superpixel $s$. Finally, the superpixel-level ED was defined as the mean L2 distance between each normalized pixel spectrum and the mean normalized spectrum:
\begin{equation}
d_s = \frac{1}{n_s}\sum_{p=1}^{n_s}
\left\|\tilde{\mathbf{f}}_p-\bar{\mathbf{f}}_s\right\|_2.
\end{equation}
Smaller $d_s$ values indicate greater spectral-shape homogeneity within the superpixel.\\

Superpixels with a $d_s$ value above the image-specific 50th percentile were excluded from the annotations for that image.\\
Last, the spectral angle mapper (SAM) describes the similarity between two spectral vectors as the angle between them in the spectral feature space. Smaller angles indicate greater spectral similarity. Because SAM depends on vector direction rather than magnitude, it is insensitive to multiplicative intensity differences. For a superpixel $s$, the mean spectral angle between its mean spectrum $\bar{\mathbf{f}}_s$ and the spectrum $\mathbf{f}_p$ of each pixel $p$ is calculated in degrees as follows:\\

\begin{equation}
SAM_s = \frac{180}{\pi n_s}\sum_{p=1}^{n_s}
\arccos\left(
\frac{\bar{\mathbf{f}}_s^{\mathsf{T}}\mathbf{f}_p}
{\left\|\bar{\mathbf{f}}_s\right\|_2
 \left\|\mathbf{f}_p\right\|_2}
\right),
\end{equation}

where $n_s$ is the number of pixels in superpixel $s$. All SAM values reported in this study are expressed in degrees. Superpixels with a $SAM_s$ value above the 50th percentile of the corresponding image are excluded from the annotations, see Figure \ref{fig:visualization} (b) and (c). A full overview of the dataset, including the number of images per patient and the number of superpixels before and after filtering for each class, is provided in the supplementary material (Table \ref{tab:data_characteristics}).\par

\subsection{HELiCoiD dataset (public)}
The public HELiCoiD setup differs from the system used to acquire the internal dataset \cite{leon_hyperspectral_2023,fabelo_-vivo_2019}. A Hyperspec VNIR A-Series pushbroom camera (Headwall Photonics Inc., Fitchburg, MA, USA) captured 826 spectral channels spanning 400--1000 nm with a spectral resolution of 2--3 nm. Table \ref{tab:acquisition_params} summarizes the main setup characteristics. The public dataset was acquired during three clinical-study campaigns and contains 61 hyperspectral images from 34 patients: 10 with LGG, 3 with HGG, 15 with GBM, and 6 with other, including secondary, brain tumors. Images were processed using dark and white references as described in the original publication \cite{leon_hyperspectral_2023}.\par 

\section{Method}
\subsection{Spectral characterization of glioma datasets}
Spectral characterization methods are applied to investigate the extent and nature of spectral differences between various types of tissue (healthy vs. tumor), data sets captured by different devices, and patient variations. The goal is to provide a comprehensive analysis of the spectral characteristics and the inherent variations both within and between these factors. This understanding can serve as a foundational step in developing a robust model for brain tumor classification, particularly for surgical resection.\par

\subsubsection{Spectral characteristics of healthy and brain tumor tissue across datasets}
Average spectra per tissue type are calculated first per patient and subsequently averaged over all patients in a dataset for both the internal and public dataset. The InVivo-LGG dataset consists of 104 spectral bands, while the HELiCoiD dataset contains 826 bands. SAM was calculated to quantify the differences between the spectral signatures of healthy and tumor pixels for each dataset. For a simple and transparent equal-dimensional comparison, we selected the first 100 bands from the InVivo-LGG dataset and paired each band with the closest wavelength in the HELiCoiD dataset, resulting in 100 matched bands from each dataset before calculating SAM. This pragmatic matching procedure was chosen to keep the descriptive comparison straightforward; it does not constitute full harmonization of the devices' spectral resolutions or response functions.\par

\subsubsection{Inter-patient versus intra-patient variability}
\begin{figure}[t!]
    \centering
    \includegraphics[width=0.95\linewidth]{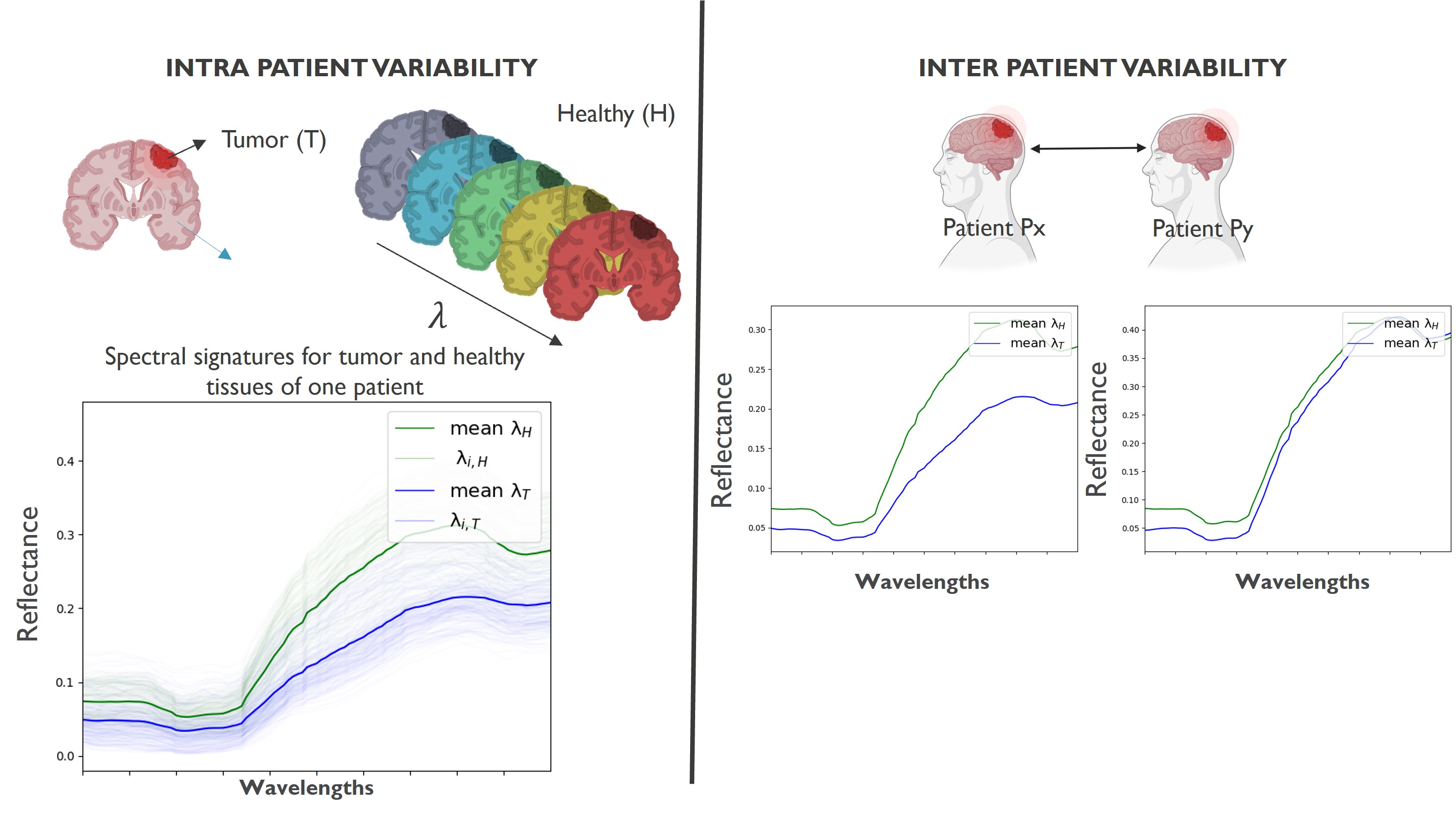}
    \caption{Visual overview for calculating the intra-patient variability (left) and inter-patient variability (right).}
    \label{fig:patient_var}
\end{figure}

The following calculations were performed separately for each dataset and tissue class $t\in\{\text{healthy},\text{tumor}\}$. Let $\mathbf{x}_{qti}\in\mathbb{R}^{B}$ denote the $B$-band spectrum of pixel $i$ from patient $q$ and tissue class $t$, and let $N_{qt}$ be the number of pixels available for that patient and class. The patient-level mean spectrum was
\begin{equation}
\boldsymbol{\mu}_{qt}=\frac{1}{N_{qt}}\sum_{i=1}^{N_{qt}}\mathbf{x}_{qti}
\end{equation}
For two spectral vectors $\mathbf{a}$ and $\mathbf{b}$, SAM in degrees was defined as
\begin{equation}
\theta(\mathbf{a},\mathbf{b})=
\frac{180}{\pi}\arccos\left(
\frac{\mathbf{a}^{\mathsf{T}}\mathbf{b}}
{\lVert\mathbf{a}\rVert_2\lVert\mathbf{b}\rVert_2}
\right).
\end{equation}
Intra-patient variability for patient $q$ and tissue class $t$ was the mean angular distance between each pixel spectrum and the corresponding patient-level mean spectrum:
\begin{equation}
\mathrm{INTRA}_{qt}=
\frac{1}{N_{qt}}\sum_{i=1}^{N_{qt}}
\theta\!\left(\mathbf{x}_{qti},\boldsymbol{\mu}_{qt}\right).
\end{equation}
Thus, pixels were weighted equally within each patient. Dataset-level intra-patient distributions were formed from the resulting patient-level values, with each patient weighted equally (Figure \ref{fig:patient_var}, left). Let $\mathcal{P}_t$ be the set of patients containing tissue class $t$ and let $M_t=|\mathcal{P}_t|$. Inter-patient variability for a unique pair of patients $q<r$ was
\begin{equation}
D_{qrt}=\theta\!\left(\boldsymbol{\mu}_{qt},\boldsymbol{\mu}_{rt}\right),
\qquad q,r\in\mathcal{P}_t.
\end{equation}
There were $C_t=\binom{M_t}{2}$ unique patient pairs for tissue class $t$. Their equally weighted mean defined the aggregate inter-patient variability:
\begin{equation}
\mathrm{INTER}_{t}=
\frac{1}{C_t}\sum_{\substack{q,r\in\mathcal{P}_t\\q<r}}D_{qrt}.
\end{equation}
Patient-level mean spectra were compared without weighting them by $N_{qt}$, and each unique patient pair contributed equally to $\mathrm{INTER}_{t}$ (Figure \ref{fig:patient_var}, right). Higher values indicate greater spectral variation within or between patients, respectively. The wavelength-matching procedure described above was applied before cross-dataset comparisons.\par 

\subsection{Classification methods}
Different classification methods were developed and benchmarked to distinguish healthy tissue from low-grade glioma tissue. Each superpixel was represented by a surrounding $40\times40\times104$ crop, with values outside the superpixel masked to zero. First, a spectral classifier was developed to associate a spectral signature with healthy or tumor tissue. The normalized mean spectrum of each superpixel, comprising 104 features, was used to train an XGBoost classifier. This method uses spectral information but no additional spatial context.\par

\begin{figure*}[h]
    \centering
    \begin{subfigure}{0.44\textwidth}
        \centering
        \includegraphics[width=\linewidth]{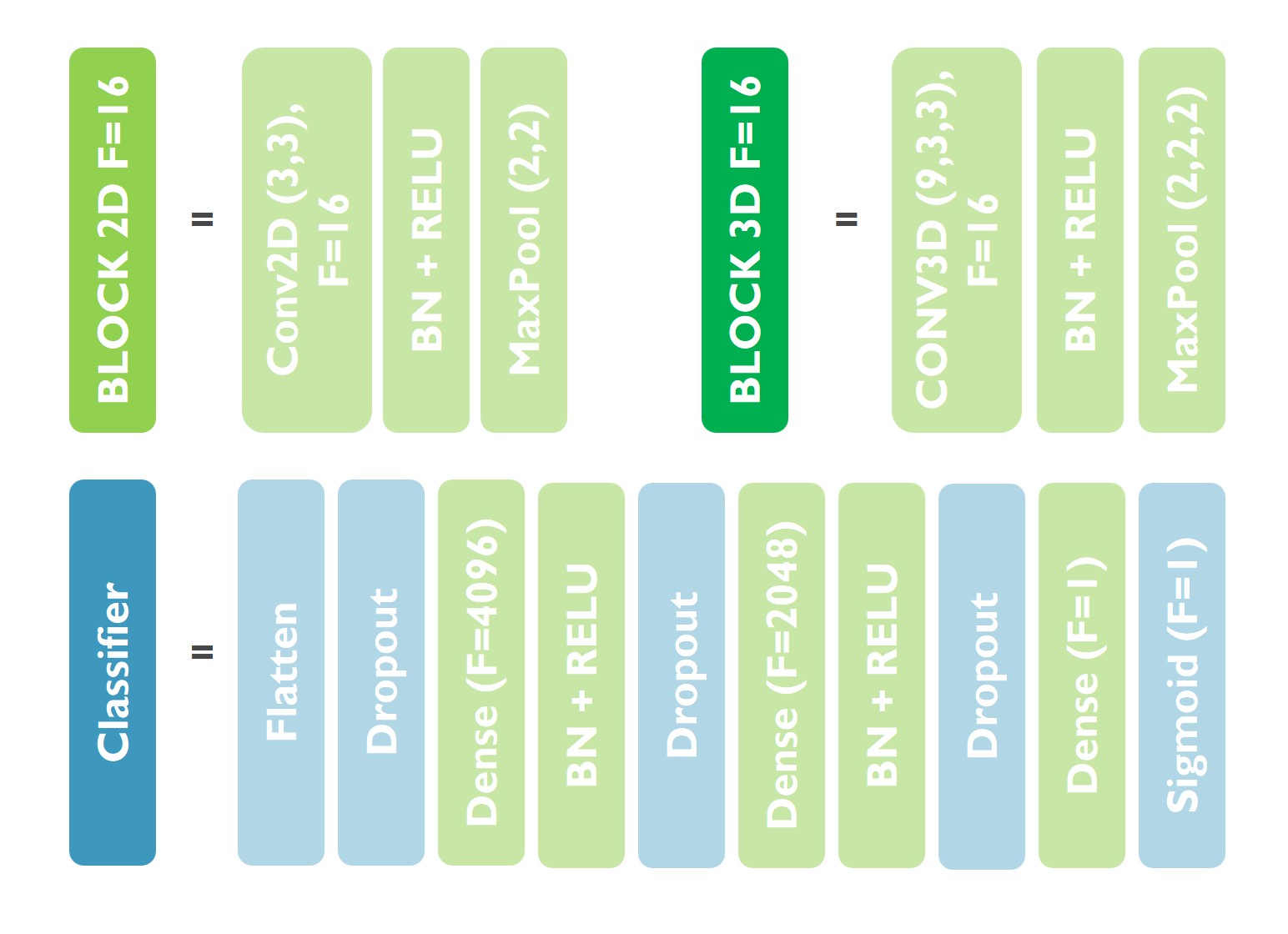}
        \caption{Neural-network building blocks}
    \end{subfigure}
    \hfill
    \begin{subfigure}{0.54\textwidth}
        \centering
        \includegraphics[width=\linewidth]{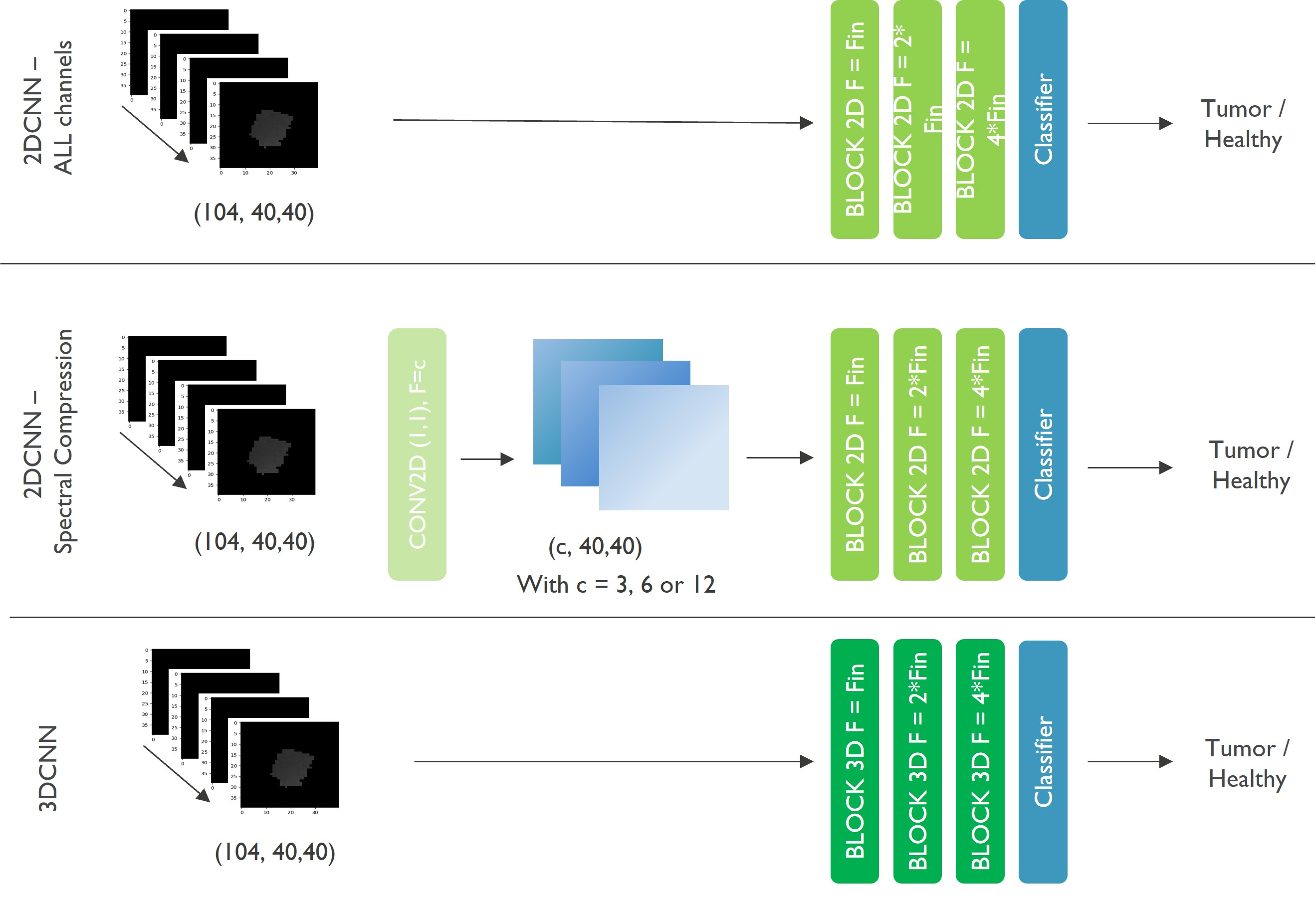}
        \caption{Neural-network architectures}
    \end{subfigure}
    \caption{Overview of (a) individual neural-network building blocks and (b) the evaluated neural-network architectures.}
    \label{fig:architectures}
\end{figure*}

Subsequently, the superpixel cube, consisting of a group of neighboring spectra, is processed as a whole, thus allowing for spectral analysis in a local spatial neighborhood. Methods such as 2D and 3D CNNs can capture variability of spectra along this cube and thus incorporate spatial information in addition to spectral information. An overview of the different methods is visualized in Figure \ref{fig:architectures}.\par 

The two-dimensional CNN using all bands (2D CNN--All channels) consists of 2D kernels learning spatial correlations along the full channel (spectral) dimension. The CNN contains 3 alternating blocks consisting of convolutional layers with a kernel size of (3x3), batch normalization, ReLU activation function and a max pooling layer with a kernel size of (2x2). At each iteration of this block, the number of feature maps is doubled. These 3 blocks are followed by two fully connected layers and a last classification layer with sigmoid activation function for final classification.\par 

A second 2D CNN, similar to the first, is implemented with an additional spectral compression layer at the beginning. A convolutional layer with a kernel size of 1x1 is used to reduce the input to 3 or 12 channels before entering a similar 2D CNN as explained above. These models are referred to as 2D-CNN-ch3 and 2D-CNN-ch12.\par

Three-dimensional CNNs consist of 3D kernels learning a mix of spatial and spectral features in a local neighbourhood. The CNN contains 3 alternating blocks consisting of convolutional layers with a kernel size of (9x3x3), batch-normalization, ReLU activation function and a max pooling layer with a kernel size of (2x2x2). At each iteration of this block, the amount of feature maps is doubled. These 3 blocks are followed by two fully connected layers and a last classification layer with sigmoid activation function for final classification. This model is further in the text referred to as 3D CNN.\par

\section{Experimental design}
\subsection{Study design}
\begin{figure*}
    \centering
    \includegraphics[width=1\textwidth]{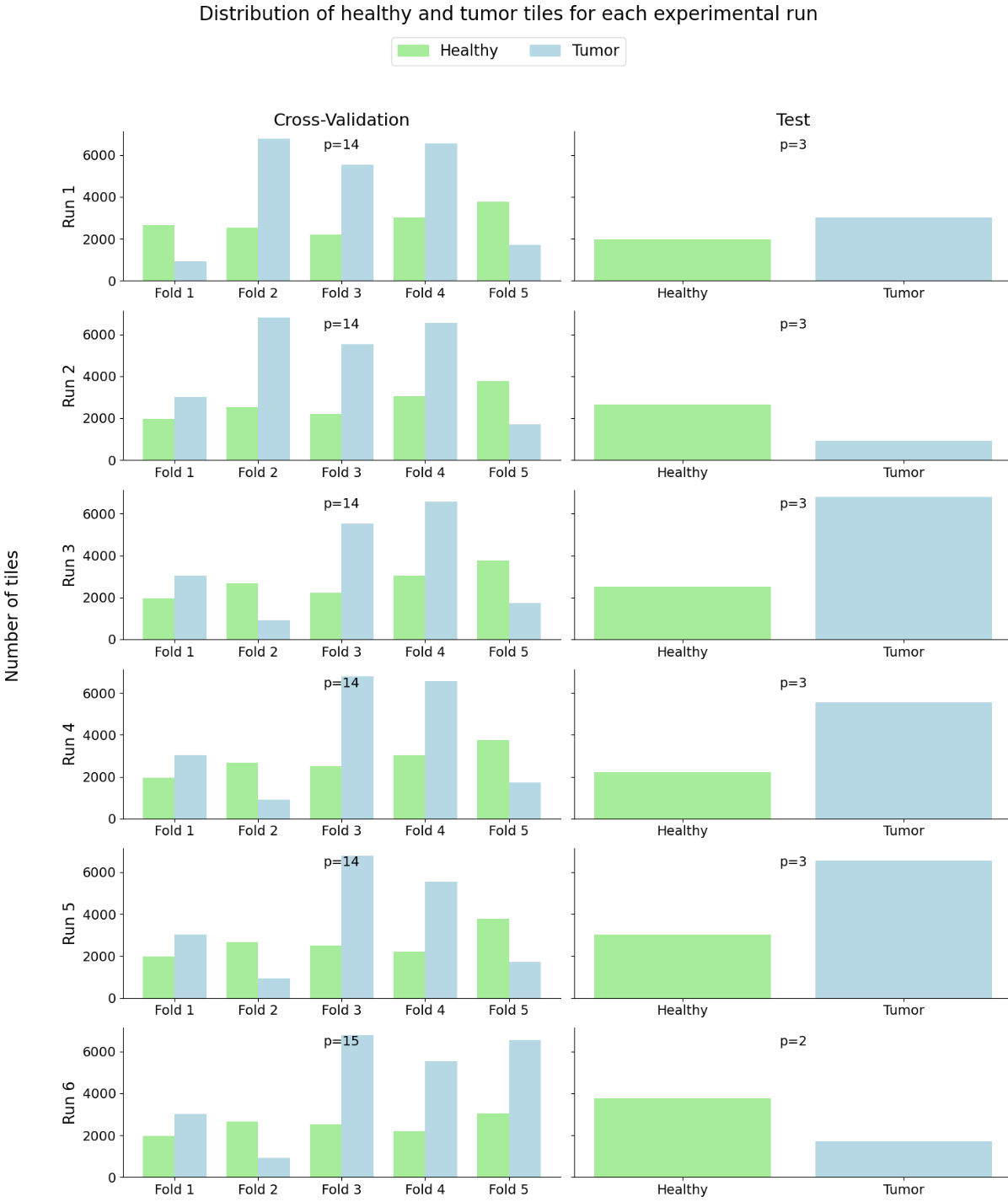}
    \caption{Overview of the patient and superpixel distribution across the six outer folds of the nested patient-level cross-validation design. Each fold is used once as the independent outer test set, while the remaining five folds are used for inner cross-validation and model optimization.}
    \label{fig:datadistribution}
\end{figure*}

To develop and evaluate the ML/DL methods explained in the previous section, we used nested, patient-level cross-validation on the 17-patient modeling cohort after excluding patient 10. The 16 patients containing both healthy and LGG annotations were divided once into six outer folds at the patient level, such that all images and superpixels from a patient remained in the same fold. The same outer-fold assignments were used for every HSI- and RGB-based model to enable paired modality comparisons. Each outer fold served once as the independent test set, resulting in six outer runs and ensuring that every two-class patient was evaluated exactly once. Patient 3, who contained only healthy-tissue annotations, was restricted to model development and was never included in an outer test fold. Figure \ref{fig:datadistribution} shows the patient and superpixel distribution across the six outer folds.\par

Within each outer run, the five non-test folds were used for five-fold inner cross-validation and hyperparameter optimization, including batch size, number of features, and learning rate. Hyperparameters were selected independently within each outer run; consequently, the selected configuration could differ between outer runs. For the selected configuration, the five inner-fold models were applied to the held-out outer test fold, and their predicted probabilities were averaged. This cross-validated ensemble reduces dependence on any single inner split while preserving complete separation of the outer test patients from model and hyperparameter selection. The purpose of this design was not to identify one universally best architecture, but to compare how effectively models using RGB and HSI inputs extract information under identical patient splits and evaluation conditions.\par

\subsection{Experiments}
Several AI architectures were optimized and benchmarked using HSI data. To compare HSI- and RGB-based performance, a 2D CNN was also developed for RGB input and optimized using the same patient-level data splits as the HSI-based models.

\subsection{Training details}
All models were trained for 50 epochs using binary cross-entropy loss and the Adam optimizer. A learning-rate scheduler reduced the learning rate by a factor of 2 every 10 epochs. During each epoch, a balanced sampler selected 100 superpixels per class and patient for patients containing both labels, ensuring equal sampling of healthy and tumor tissue across patients. Patient 3 contributed only its available healthy-tissue superpixels.\par

\subsection{Evaluation}

Model performance was evaluated using F1 score, balanced accuracy (BACC), accuracy (ACC), precision (PREC), recall (REC), and ROC-AUC. Metric definitions are provided in the supplementary material (Section \ref{sup:metrics}). ROC-AUC was computed from continuous predicted probabilities before thresholding. For F1, BACC, ACC, PREC, and REC, probabilities greater than or equal to 0.5 were classified as LGG, while probabilities below 0.5 were classified as healthy tissue. Metrics were calculated separately for each eligible test patient using all retained superpixels. For each metric, every HSI-based model was compared separately with 2DCNN-RGB using a paired Wilcoxon signed-rank test. The paired unit was the patient ($n=16$): each pair comprised the metric value from an HSI-based model and the corresponding value from 2DCNN-RGB for the same patient. Because each patient appeared in exactly one held-out outer test fold, each patient contributed one paired observation to a given model--metric comparison. A difference was considered statistically significant when $p<0.05$.\par

\section{Results}
\subsection{Spectral characterization}
\begin{figure}
    \centering
    \includegraphics[width=1\linewidth]{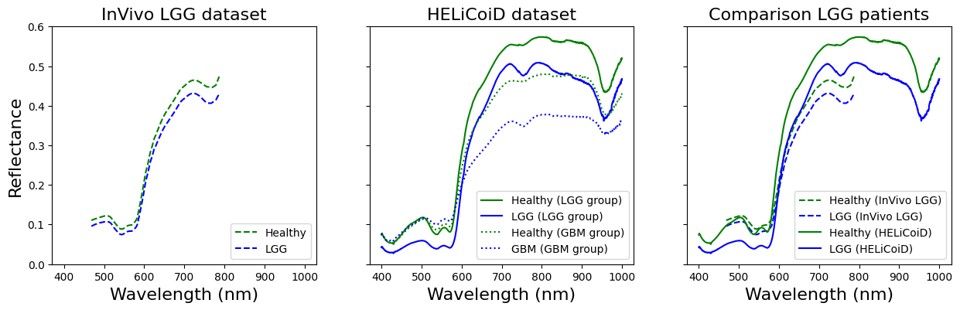}
    \caption{Mean spectra of healthy (green) and tumor (blue) tissue in the internal InVivo-LGG dataset and the public HELiCoiD dataset.}
    \label{fig:avg_reflectance}
\end{figure}

Figure \ref{fig:avg_reflectance} shows the mean reflectance spectra of healthy and tumor tissue across patients in both datasets. Although the datasets were acquired with different HSI systems and have different spectral resolutions and ranges, they share several spectral features. Peaks and troughs occur at similar wavelengths, including a peak near 505 nm, features near 560 nm, and a trough near 720 nm. These features may be related to the absorption characteristics of deoxyhemoglobin (Hb) and oxyhemoglobin (HbO$_{2}$). Near the isosbestic point at 584 nm, where Hb and HbO$_{2}$ have equal absorption, the healthy- and tumor-tissue spectra are similar \cite{noauthor_optical_nodate}. In both datasets, tumor tissue has lower reflectance than healthy tissue across the measured range. The HELiCoiD data also show a tumor-associated trough near 720 nm that is less prominent in healthy tissue. HELiCoiD LGG and GBM patients have SAM values of 4.9$^{\circ}$ and 4.2$^{\circ}$, respectively, indicating differences between healthy- and tumor-tissue signatures. The corresponding signatures in InVivo-LGG are more similar (SAM = 0.9$^{\circ}$), which may make classification more difficult.\par
\begin{figure}
    \centering
    \includegraphics[width=1\linewidth]{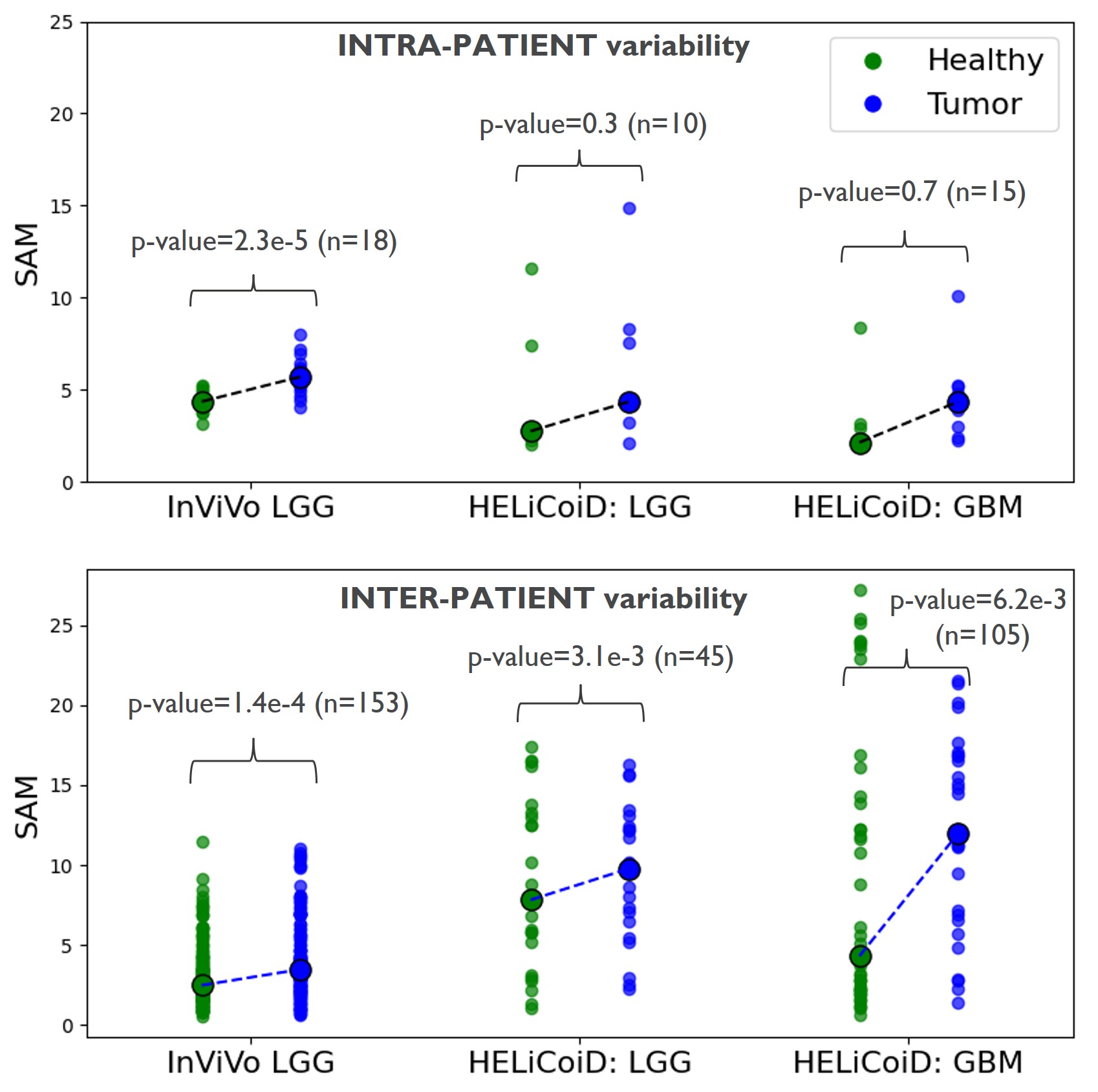}
    \caption{SAM values in degrees for (a) intra-patient variability and (b) inter-patient variability. In (a), each dot represents a patient, with green dots indicating healthy tissue and blue dots indicating tumor tissue; highlighted circles represent median values. In (b), each dot represents a pair of patients.}
    \label{fig:patientvariabilityR}
\end{figure}

 
For intra-patient variability (Figure \ref{fig:patientvariabilityR}), tumor tissue has higher SAM values than healthy tissue in all datasets, with a statistically significant difference reported only for InVivo-LGG. This finding indicates greater within-patient spectral variability in tumor tissue. Mean inter-patient variability is higher in HELiCoiD than in InVivo-LGG for both tissue classes. Across datasets, tumor tissue has higher inter-patient SAM values than healthy tissue ($p<0.05$).\par

\subsection{Classification results}
Table \ref{tab:model_results} summarizes the performance of all models, calculated per patient over the superpixels from the test cohorts across the different runs. Among the HSI-based models, XGBoost achieved the highest median F1 score, accuracy, and recall; the 3D CNN achieved the highest median balanced accuracy and ROC-AUC; and 2DCNN-ch3 achieved the highest median precision. These numerical differences between HSI-based model families were not statistically significant. Overall, the HSI-based models performed significantly better than the RGB-based model, demonstrating the added value of hyperspectral information for this classification task.\par

\begin{table}[h]
\caption{Test-set classification metrics summarized per patient as the median $\pm$ 95\% confidence interval. Boldface indicates a statistically significant paired difference between that HSI-based model and 2DCNN-RGB for the corresponding metric. No statistically significant differences were observed among the HSI-based model families.}
\centering
\scalebox{0.7}{
\begin{tabular}{p{2.5cm}|c|ccccccc}
\hline
Models& \#patients & F1 & ACC & b. ACC & AUC-ROC & PREC & RECALL\\
\hline
\hline
XGB	& 16	& \textbf{0.68 $\pm$ 0.09}& \textbf{0.68 $\pm$ 0.06}&	\textbf{0.68 $\pm$ 0.06}&	\textbf{0.68 $\pm$ 0.06}&	0.68 $\pm$ 0.13&	\textbf{0.74 $\pm$ 0.08}\\
2DCNN-All & 16	&	0.60 $\pm$ 0.11&	0.64 $\pm$ 0.07&	\textbf{0.67 $\pm$ 0.06}&	\textbf{0.73 $\pm$ 0.07}&	0.67 $\pm$ 0.15&	0.66 $\pm$ 0.15\\
2DCNN-ch12 & 16	&	0.58 $\pm$ 0.13&	0.63 $\pm$ 0.09&	\textbf{0.67 $\pm$ 0.08}&	\textbf{0.74 $\pm$ 0.09}&	0.68 $\pm$ 0.15&	0.64 $\pm$ 0.16\\
2DCNN-ch3 & 16	&	0.59 $\pm$ 0.11&	0.63 $\pm$ 0.08&	0.67 $\pm$ 0.08&	0.74 $\pm$ 0.10&	0.69 $\pm$ 0.13&	0.63 $\pm$ 0.16\\
3DCNN & 16	&	0.64 $\pm$ 0.10&	0.66 $\pm$ 0.08&	\textbf{0.69 $\pm$ 0.07}&	\textbf{0.75 $\pm$ 0.08}	&0.69 $\pm$ 0.13	&0.70 $\pm$ 0.14\\
\hline
2DCNN-RGB& 16&		0.56 $\pm$ 0.09&	0.59 $\pm$ 0.05&	0.59 $\pm$ 0.06&	0.64 $\pm$ 0.08&	0.63 $\pm$ 0.13&	0.55 $\pm$ 0.10\\

\hline
\end{tabular}}
\label{tab:model_results}
\end{table}

\begin{figure}
    \centering
    \includegraphics[width=0.7\linewidth]{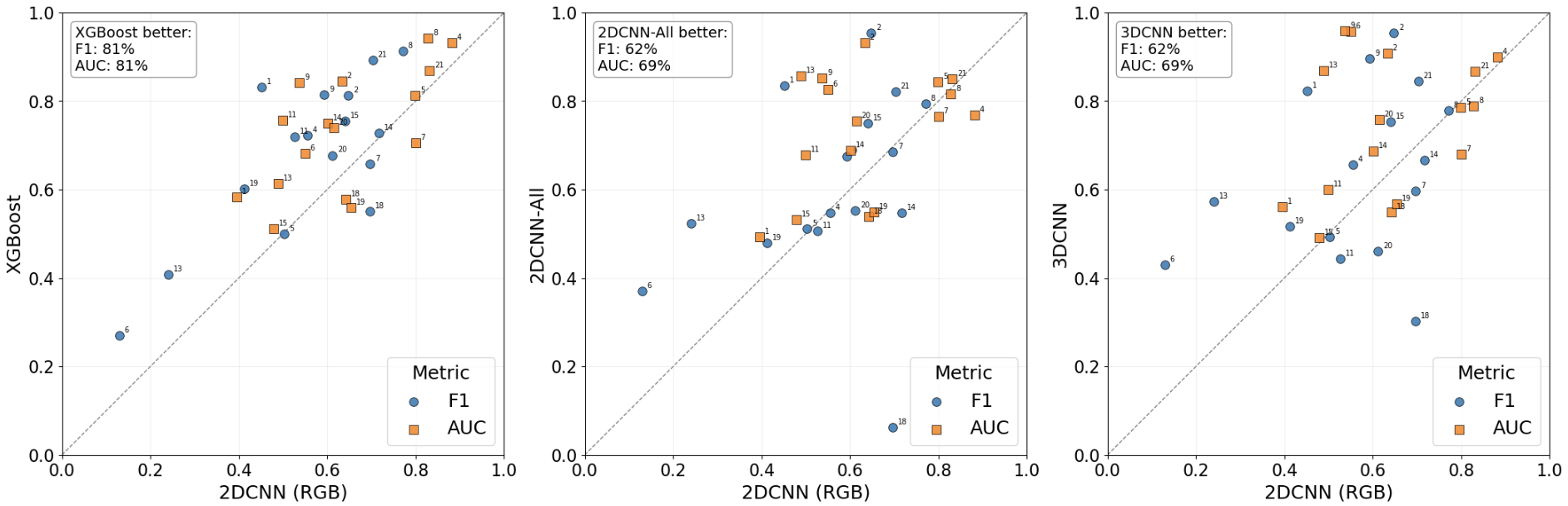}
    \caption{Scatterplot of patient-level F1 and AUC-ROC values obtained from the six outer test folds.}
    \label{fig:scatterplots}
\end{figure}

\begin{figure}
    \centering
    \includegraphics[width=0.7\linewidth]{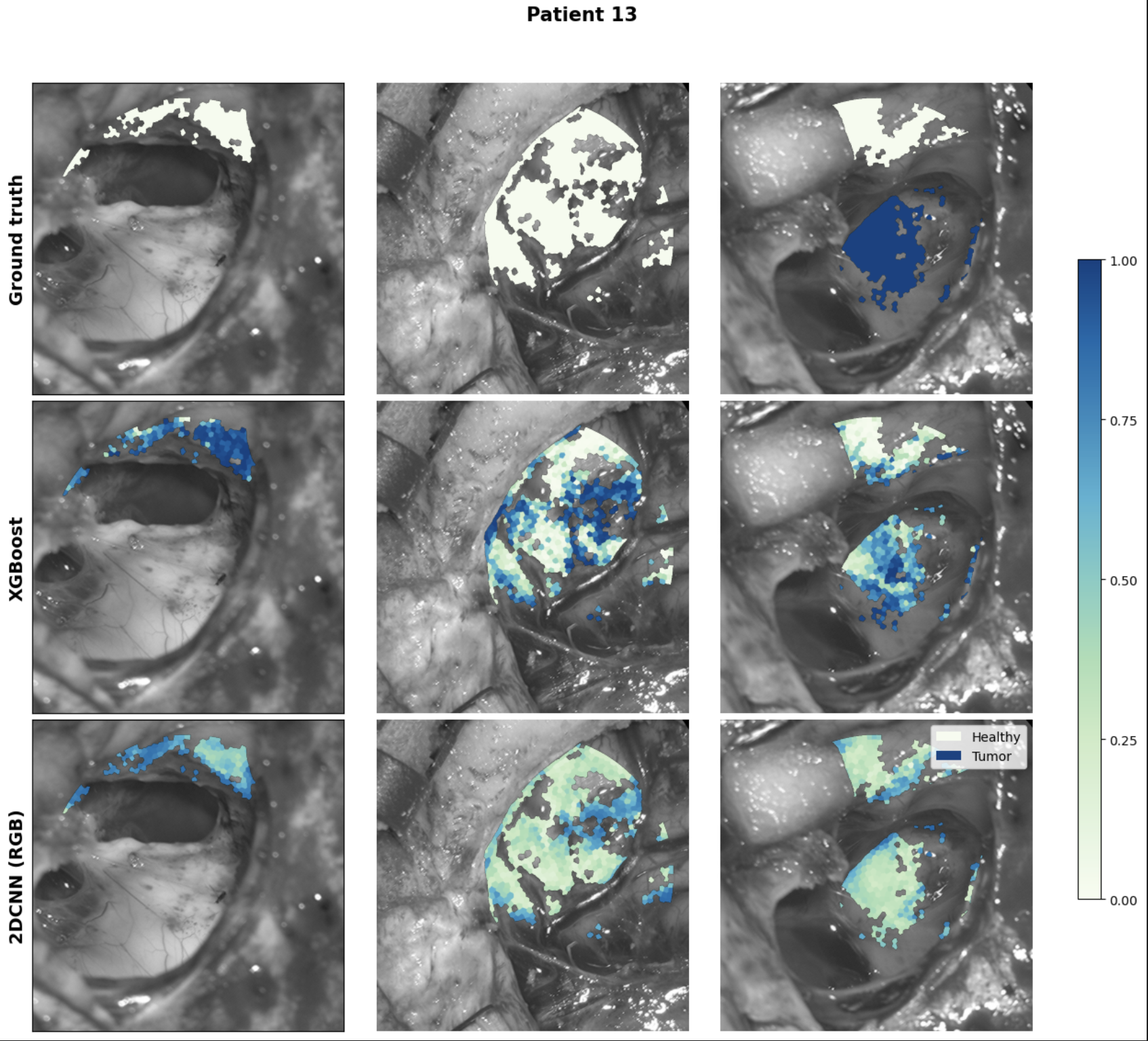}
    \caption{Visual examples of patient 13}
    \label{fig:ex1}
\end{figure}

\begin{figure}
    \centering
    \includegraphics[width=0.7\linewidth]{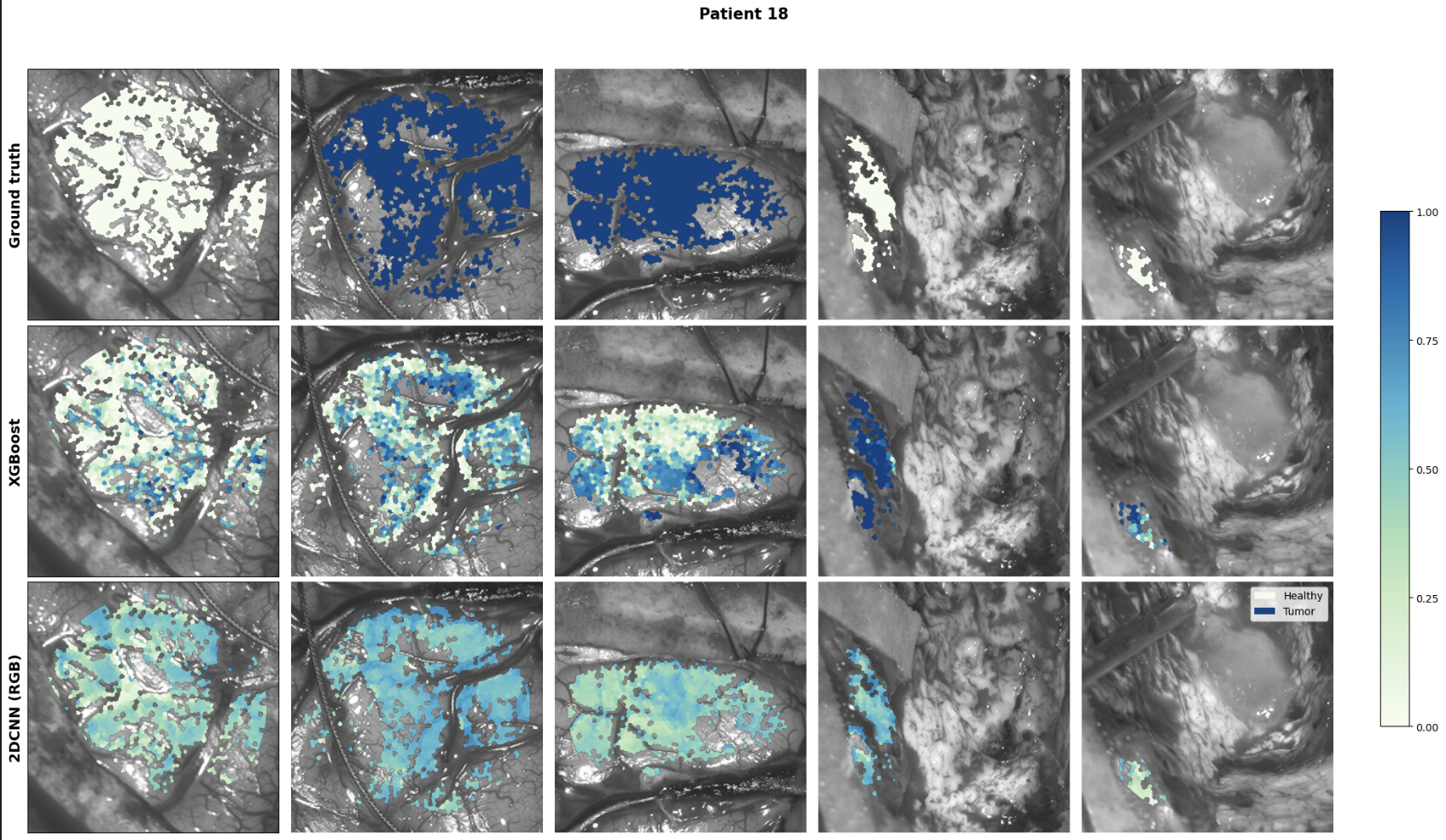}
    \caption{Visual examples of patient 18}
    \label{fig:ex2}
\end{figure}

\begin{figure}[h]
    \centering
    \includegraphics[width=0.95\linewidth]{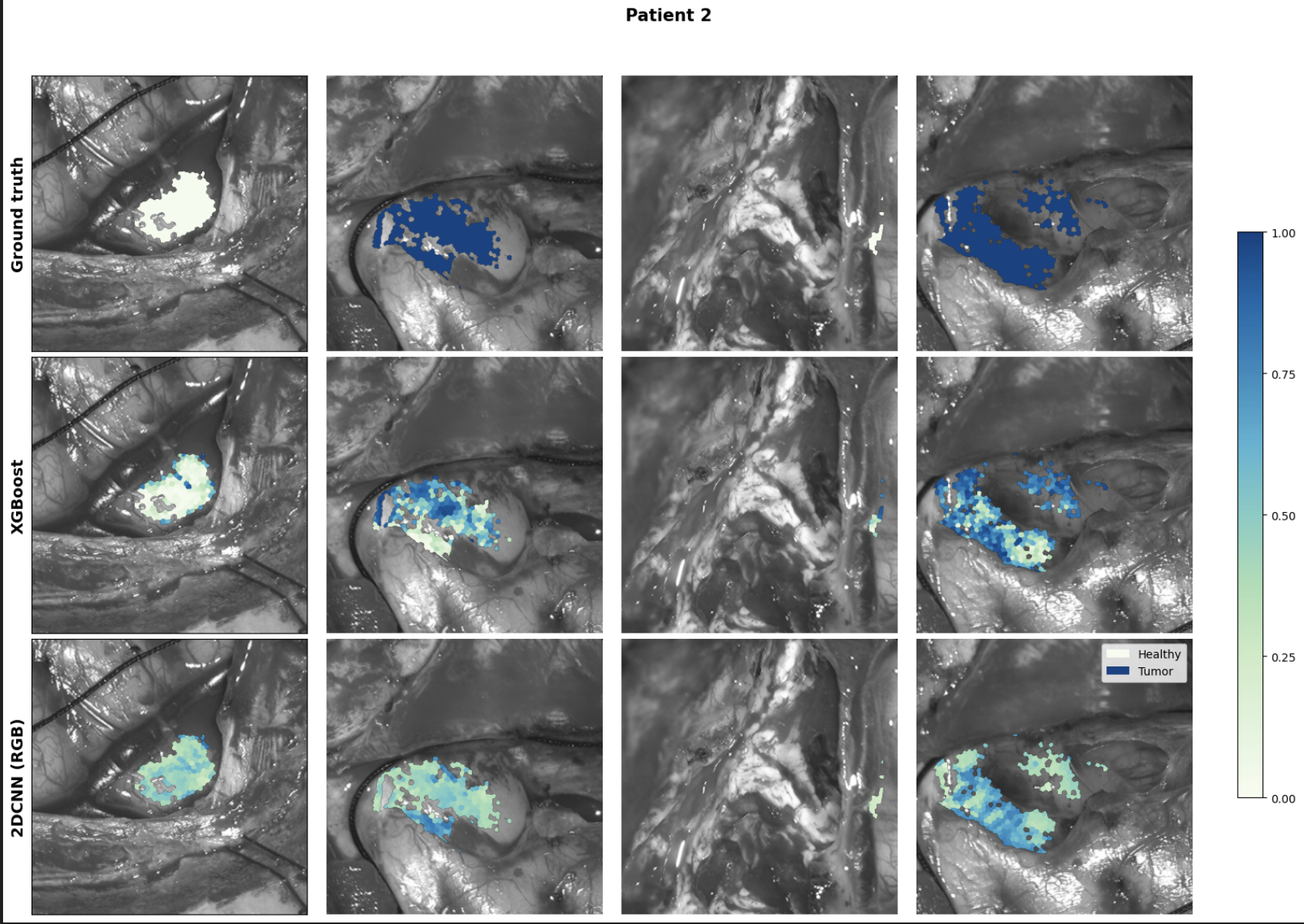}
    \caption{Visual examples of patient 2}
    \label{fig:ex3}
\end{figure}

Figure \ref{fig:scatterplots} compares patient-level F1 and AUC-ROC scores for the RGB model on the x-axis and the XGB, 2DCNN-All and 3D CNN on the y-axis. The HSI-based models achieve higher performances for most of the patients. Predictions from the XGBboost and RGB model are shown in Figures \ref{fig:ex1}, \ref{fig:ex2}, and \ref{fig:ex3}.\par

\section{Discussion}
This study extends the intraoperative brain HSI literature in two specific ways. First, it reports a clinical cohort focused on low-grade glioma, complementing prior studies that predominantly evaluated high-grade glioma, glioblastoma, or mixed-grade cohorts. Second, it examines the HSI--RGB difference through paired acquisitions, matched tissue locations, identical patient-level splits, and a common evaluation procedure. Spectral characterization of two datasets acquired with different clinical HSI systems also revealed similar features in the shapes of their reflectance curves despite differences in spatial and spectral resolution.\par

Measured tissue reflectance is influenced by hemoglobin absorption, blood volume, and oxygenation in the visible and near-infrared range \cite{noauthor_optical_nodate}. Accordingly, features associated with Hb and HbO$_{2}$ are present in both InVivo-LGG and HELiCoiD. LGG tissue shows lower reflectance than healthy brain tissue, particularly toward the near-infrared. Differences in tissue water content, edema, and vascularity may contribute to this observation \cite{Bartha464,Law2003,Law2006,Essig2013}, but these properties were not measured directly in this study. Microvascular proliferation is also more pronounced in high-grade gliomas \cite{Provenzale2002,Wong2012}. These biological explanations should therefore be considered hypotheses rather than demonstrated mechanisms.\par

HELiCoiD LGG spectra show a trough near 720 nm that could be associated with deoxyhemoglobin absorption. Although this feature might be consistent with localized hypoxia \cite{Evans2004,Shengchao2021}, no physiological measurements were available to validate that interpretation. Its absence from InVivo-LGG could instead reflect differences in spectral resolution, instrument response, illumination, tissue geometry, preprocessing, clinical conditions or more rigid patient diagnosis according to the WHO 2021 classification.\par

Non-biological factors also influence reflectance spectra. Ambient conditions may vary during acquisition, and the larger HELiCoiD pushbroom system may be more sensitive to environmental fluctuations than the imec Snapscan VNIR system. Conversely, HELiCoiD has higher spectral resolution and more densely sampled signatures, whereas InVivo-LGG has broader spectral response functions. Differences in spectral and spatial resolution, lenses, illumination, instrument response, acquisition protocol, and ground-truth procedures all complicate direct comparison between the datasets.\par

Reliable ground truth remains challenging in medical imaging. HELiCoiD used semi-automatic labeling: experts selected image locations with clear healthy-tissue, vessel, or tumor labels, after which pixels with similar spectral characteristics were assigned to those groups. This procedure increases separation between the resulting classes. In contrast, the present study broadly delineated regions using preoperative imaging and histological information from selected locations and then removed outliers, which may reduce interclass variation. Ideally, all imaging information would be accurately registered to the intraoperative frame of reference and validated against spatially matched histopathology, particularly at tumor margins.\par

Moreover, the study benchmarks several ML and DL model families. No statistically significant differences were observed among the HSI-based model families, indicating that no single architecture can be considered uniformly superior in this cohort. Instead, the architectures offer complementary benefits. XGBoost processes the complete mean spectrum of a superpixel and can model nonlinear interactions between spectrally distant wavelengths, but it does not use spatial context. In contrast, the 3D CNN jointly learns local spectral and spatial correlations within a superpixel cube. Under the paired evaluation design, the HSI-based models performed significantly better overall than the RGB-based model, supporting the added value of hyperspectral information for this low-grade glioma classification task while remaining specific to the selected tissue regions and cohort studied here. Previous studies have demonstrated the relevance of HSI for brain tumor detection, particularly for glioblastoma \cite{leon_hyperspectral_2023, fabelo_-vivo_2019, manni_hyperspectral_2020}, while related work outside medicine has explored combined spectral and spatial modeling \cite{atik_dual-stream_2024, dong_deep_2019, kumar_deep_2024, li_deep_2019, shenming_new_2022}. A transformer-based architecture could combine long-range spectral interactions with spatial context more directly; however, such models generally require substantially more training data than are available in the present cohort and should therefore be investigated in larger studies. Czempiel et al. \cite{czempiel2024rgb} reconstructed spectral images from RGB images using deep learning and the public HeiPorSPECTRAL pig dataset \cite{studier2023heiporspectral}. This finding suggests that RGB images may contain information useful for spectral reconstruction. The present study evaluated direct RGB classification and did not investigate a two-stage RGB-to-HSI reconstruction and classification pipeline.
\par

\section{Limitations}
This study has several limitations. A single surgeon provided the ground-truth delineations of tumor and healthy-tissue regions, informed by pathology at selected locations. No inter-rater or intra-rater agreement measurements were obtained. The classification task included only healthy and tumor tissue and excluded vessels and other background classes that would be important in clinical practice. Future studies should include these classes and explicitly evaluate specular reflections and small vessels.\par

An important limitation of the RGB--HSI comparison is that sample definition and filtering were based on the HSI data. SLIC superpixels were generated from the hyperspectral images, and intensity, ED, and SAM thresholds were subsequently used to retain spectrally homogeneous regions. This procedure removes ambiguous and spectrally heterogeneous observations before model evaluation and may therefore favor HSI-based models. Evaluating the RGB models at the same retained spatial locations ensures a paired comparison but does not eliminate this selection bias, because the inclusion criteria were still determined from HSI. Consequently, the observed performance difference between HSI and RGB should be interpreted in the context of this HSI-derived sample selection. Future work should quantify the effect through ablation analyses without spectral filtering or define evaluation regions using a modality-independent procedure.\par

The filtering procedure also excludes mixed-label superpixels at tissue boundaries, vessels, blood-contaminated regions, specular reflections, very dark regions, and spectrally heterogeneous tissue. These conditions are common in the intraoperative field, and mixed or heterogeneous regions may include tumor margins that are particularly relevant for surgical guidance. Therefore, the present results demonstrate classification performance for selected, spectrally homogeneous tissue regions and should not be interpreted as evidence of general intraoperative tumor detection or reliable margin delineation under all surgical conditions. Future studies should explicitly evaluate these excluded conditions and tumor-boundary regions.\par

In addition, filtering may affect the spectral characterization itself. Excluding observations outside the image-specific 5th--95th intensity percentiles and subsequently removing superpixels above the image-specific 50th percentiles for ED and SAM can alter estimated tissue reflectance distributions and reduce measured within-class variability. The supplementary material reports the total number of superpixels before and after filtering, but it does not quantify retention after each individual filtering step. Future analyses should report stepwise sample retention and repeat the spectral comparisons before and after filtering to assess the sensitivity of the observed healthy--tumor differences to these choices.\par

Finally, the extent of spatial information used by the neural networks is limited because only superpixels are classified. Broadening the scope towards segmentation-based approaches that include both local and global spatial information in addition to spectral information is an important direction for future work.\par

\section{Conclusion}
This clinical study focuses on low-grade glioma, complementing intraoperative HSI studies centered on high-grade glioma, glioblastoma, or mixed-grade cohorts. It also provides a paired HSI--RGB comparison using matched tissue locations and identical patient-level evaluation splits. Despite differences between the two HSI systems, both datasets show similar spectral features in brain tissue, and cross-dataset characterization indicates greater inter-patient spectral variability in tumor tissue than in healthy tissue. The 3D CNN, which evaluates spectral signatures within a local spatial context, achieved a median ROC-AUC of 74.5\% and a median F1 score of 63.7\%, compared with 64.0\% and 55.6\%, respectively, for the RGB-based model. These findings quantify the HSI--RGB performance difference for selected tissue regions in this low-grade glioma cohort.

\section{Acknowledgments}
This work has received funding from the Flemish Government under the ``Onderzoeksprogramma Artificiele Intelligentie (AI) Vlaanderen'' programme.

\bibliographystyle{elsarticle-num} 
\bibliography{iHSI}

\newpage
\appendix

\section{Registration Examples}

\begin{figure}[h]
    \centering
    \includegraphics[width=1\linewidth]{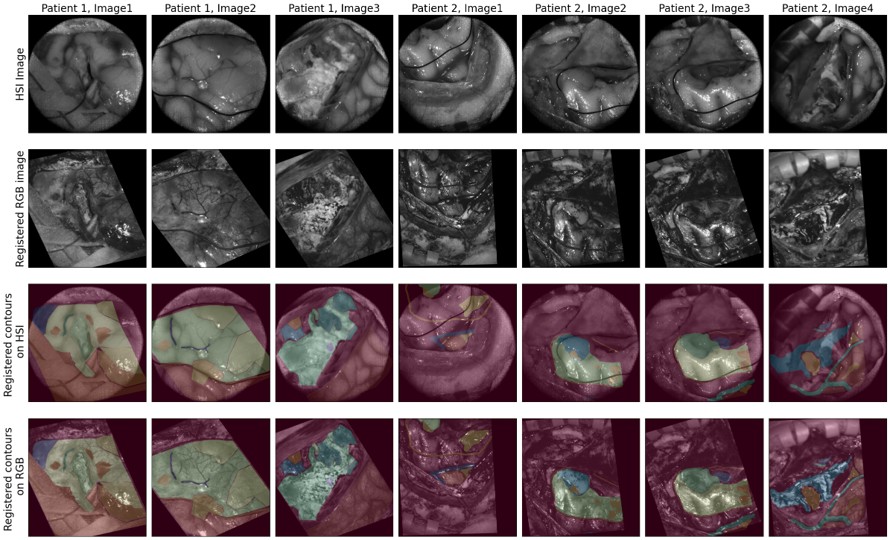}
    \caption{Examples of both good and bad registrations of RGB image and corresponding contours with HSI. Bad registrations were manually registered afterwards.}
    \label{fig:registrationexamples}
\end{figure}

\newpage
\section{Data characteristics}

\begin{table}[h]
\centering
\caption{Overview of all 18 enrolled patients, including the number of acquired images and superpixels before and after filtering. Patient 10 is shown for cohort transparency but was excluded from all model development and evaluation because fluorescence imaging was used. Patient 3 contained only healthy-tissue annotations and was used for training but excluded from patient-level binary classification evaluation.}
\label{tab:data_characteristics}
\begin{adjustbox}{max width=\textwidth}
\begin{tabular}{c|c|p{2cm}p{2cm}p{2cm}p{2cm}}
\hline
Patient & Nr. of images & Initial Superpixels (Tumor) & Filtered Superpixels (Tumor)&Initial Superpixels (Healthy) & Filtered Superpixels (Healthy)  \\
\hline
\hline
1 & 2 & 574 & 250 & 230 & 101 \\
2 & 4 & 1316 & 564 & 361 & 162 \\
3 & 2 & 0 & 0 & 588 & 274 \\
4 & 4 & 467 & 212 & 1358 & 648 \\
5 & 4 & 1321 & 441 & 3449 & 1732 \\
6 & 4 & 626 & 264 & 4129 & 1842 \\
7 & 5 & 3344 & 1456 & 4267 & 1923 \\
8 & 4 & 4761 & 2033 & 2015 & 985 \\
9 & 5 & 3390 & 1504 & 2258 & 996 \\
10 & 2 & 607 & 329 & 2262 & 889 \\
11 & 5 & 6828 & 3019 & 2433 & 1052 \\
13 & 3 & 614 & 225 & 1713 & 832 \\
14 & 7 & 10852 & 4876 & 1487 & 681 \\
15 & 4 & 2446 & 1090 & 1146 & 486 \\
18 & 5 & 3587 & 1699 & 2648 & 1186 \\
19 & 4 & 3100 & 1296 & 2191 & 1113 \\
20 & 6 & 5608 & 2462 & 3195 & 1526 \\
21 & 5 & 7317 & 3147 & 1150 & 610 \\
\hline
\end{tabular}
\end{adjustbox}
\end{table}

\newpage
\section{Binary classification metrics}
\label{sup:metrics}

\begin{table}[h]
\centering
\begin{tabular}{l| l| c c}
    &  & \multicolumn{2}{c}{Prediction}  \\
   \hline
   &  & Positive & Negative\\
   \hline
  \multirow{2}{*}{Ground truth} & Positive & True positive (TP) & False Negative (FN)\\
   & Negative &  False Positive (FP) & True Negative (TN)\\
\end{tabular}
\caption{Binary confusion matrix.}
\label{tab:confusion_matrix}
\end{table}

Accuracy is the proportion of correct predictions among all predictions:
\begin{equation}
\mathrm{ACC}=\frac{TP+TN}{TP+TN+FP+FN}.
\end{equation}

Precision is the proportion of positive predictions that are true positives:
\begin{equation}
\mathrm{PREC}=\frac{TP}{TP+FP}.
\end{equation}

Recall, also called sensitivity or the true-positive rate (TPR), is the proportion of actual positives that are correctly identified:
\begin{equation}
\mathrm{REC}=\mathrm{TPR}=\frac{TP}{TP+FN}.
\end{equation}

Specificity is the proportion of actual negatives that are correctly identified:
\begin{equation}
\mathrm{SPEC}=\frac{TN}{TN+FP}.
\end{equation}

The F1 score is the harmonic mean of precision and recall:
\begin{equation}
F1=\frac{2\,\mathrm{PREC}\,\mathrm{REC}}{\mathrm{PREC}+\mathrm{REC}}
=\frac{2TP}{2TP+FP+FN}.
\end{equation}

The false-positive rate (FPR) is defined as
\begin{equation}
\mathrm{FPR}=\frac{FP}{FP+TN}.
\end{equation}
The area under the receiver-operating-characteristic curve (ROC-AUC) is
\begin{equation}
\mathrm{AUC}_{\mathrm{ROC}}=\int_0^1 \mathrm{TPR}(u)\,du,
\end{equation}
where $u$ denotes the false-positive rate as the decision threshold varies.

\newpage
\section{Results cross-validations}

\begin{figure}
    \centering
    \includegraphics[width=0.5\linewidth]{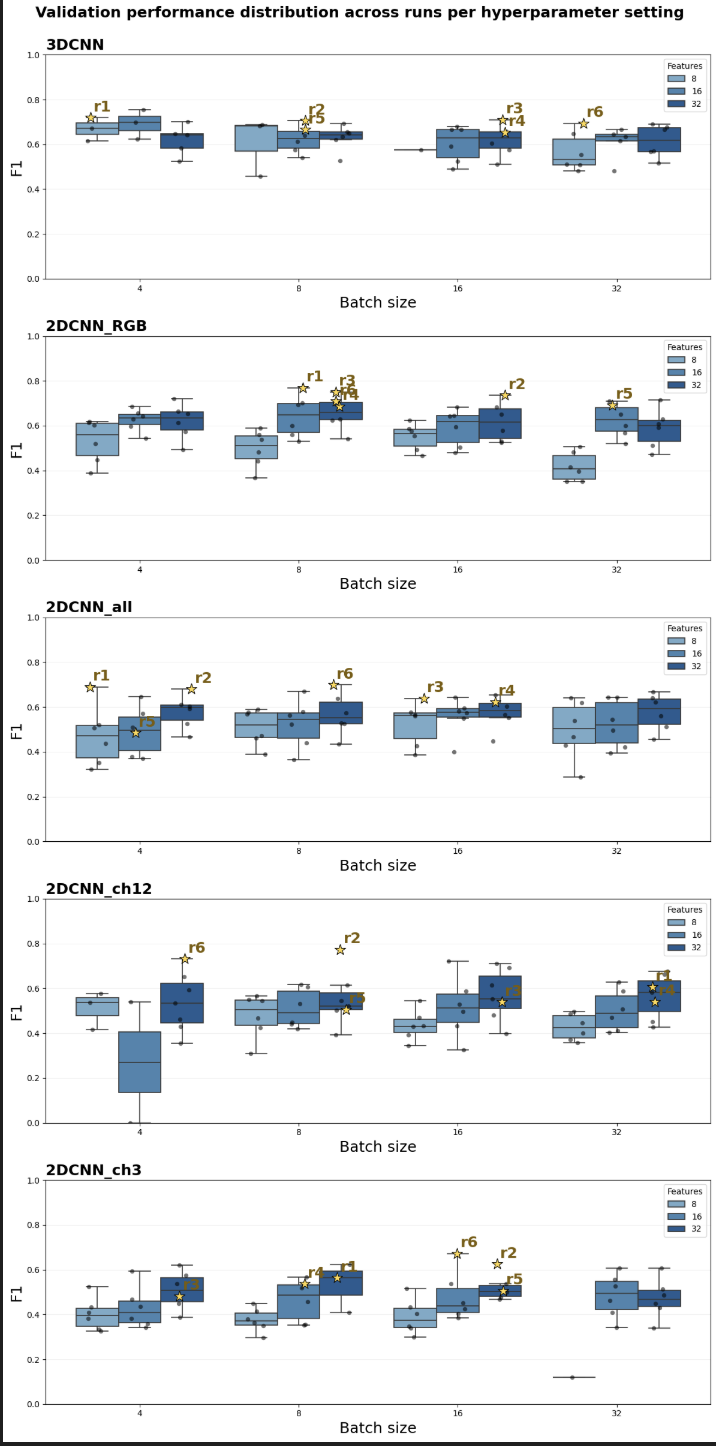}
    \caption{Hyperparameter optimization results for the cross-validations for all runs. Each sample in the boxplot represents the cross-validation score for a run given the combination of hyperparameters}
    \label{fig:xval}
\end{figure}

The results of the hyperparameter optimization using a cross-validation for each run, is visualized in Figure \ref{fig:xval}

\newpage

\end{document}